\documentclass[conference]{IEEEtran}
\IEEEoverridecommandlockouts
\usepackage{cite}
\usepackage{amsmath,amssymb,amsfonts}
\usepackage{graphicx}
\usepackage{textcomp}
\usepackage{color}
\usepackage{algorithm}
\usepackage{algpseudocode}
\usepackage{bm}
\usepackage{epstopdf}
\usepackage{multirow}
\def\BibTeX{{\rm B\kern-.05em{\sc i\kern-.025em b}\kern-.08em
    T\kern-.1667em\lower.7ex\hbox{E}\kern-.125emX}}
\begin{document}

\title{Channel Attention and Multi-level Features Fusion for Single Image Super-Resolution
}

\author{
\IEEEauthorblockN{1\textsuperscript{st} Yue Lu}
\IEEEauthorblockA{
\textit{Beijing University of} \\
\textit{Posts and Telecommunications}\\
Beijing, China \\
lu\_yue@bupt.edu.cn}
\and
\IEEEauthorblockN{2\textsuperscript{nd} Yun Zhou}
\IEEEauthorblockA{
\textit{Academy of Broadcasting Science}\\
Beijing, China \\
zhouyung@abs.ac.cn}
\and
\IEEEauthorblockN{3\textsuperscript{rd} Zhuqing Jiang}
\IEEEauthorblockA{
\textit{Beijing University of}\\
 \textit{Posts and Telecommunications}\\
Beijing, China \\
jiangzhuqing@bupt.edu.cn}
\and
\IEEEauthorblockN{4\textsuperscript{th} Xiaoqiang Guo}
\IEEEauthorblockA{
\textit{Academy of Broadcasting Science}\\
Beijing, China \\
guoxiaoqiang@abs.ac.cn}
\and
\IEEEauthorblockN{5\textsuperscript{th} Zixuan Yang}
\IEEEauthorblockA{
\textit{Beijing University of}\\
 \textit{Posts and Telecommunications}\\
Beijing, China \\
hsuanr@qq.com}

}

\maketitle

\begin{abstract}
Convolutional neural networks (CNNs) have demonstrated superior performance in super-resolution (SR). However, most CNN-based SR methods neglect the different importance among feature channels or fail to take full advantage of the hierarchical features. To address these issues, this paper presents a novel recursive unit. Firstly, at the beginning of each unit, we adopt a compact channel attention mechanism to adaptively recalibrate the channel importance of input features. Then, the multi-level features, rather than only deep-level features, are extracted and fused. Additionally, we find that it will force our model to learn more details by using the learnable upsampling method (i.e., transposed convolution) only on residual branch (instead of using it both on residual branch and identity branch) while using the bicubic interpolation on the other branch. Analytic experiments show that our method achieves competitive results compared with the state-of-the-art methods and maintains faster speed as well.
\end{abstract}


\section{Introduction}
The aim of Single Image Super-Resolution (SISR) \cite{1} is to recover a high resolution (HR) image from its corresponding low resolution (LR) input image. SISR is widely used in computer vision applications such as security and surveillance imaging, satellite imaging and medical imaging.

Since deep learning has fundamentally changed how computers learn features, the field of SISR makes impressive improvements by using Convolutional Neural Networks in recent years. Dong et al. \cite{2} firstly proposed a fully convolutional neural network, termed SRCNN, which applies SR after the bicubic interpolation operation. To improve the representational power of SR network, Kim et al. \cite{3} successfully trained a 20 layers model (VDSR) to learn the global residual instead of the actual whole image, achieving vast improvements over SRCNN. Kim et al. \cite{4} further proposed a deeply-recursive convolutional network (DRCN) to efficiently reuse weight parameters while exploiting a large image context. Lai et al. \cite{5} proposed LapSRN to progressively predict residual image in a coarse-to-fine manner with Charbonnier loss, striking a balance between reconstruction accuracy and execution time. In DRRN, Tai et al. \cite{6} built a very deep network (up to 52 layers) with deep recursions for SR.

\begin{figure*}[t]
\centering

\includegraphics[width=18cm]{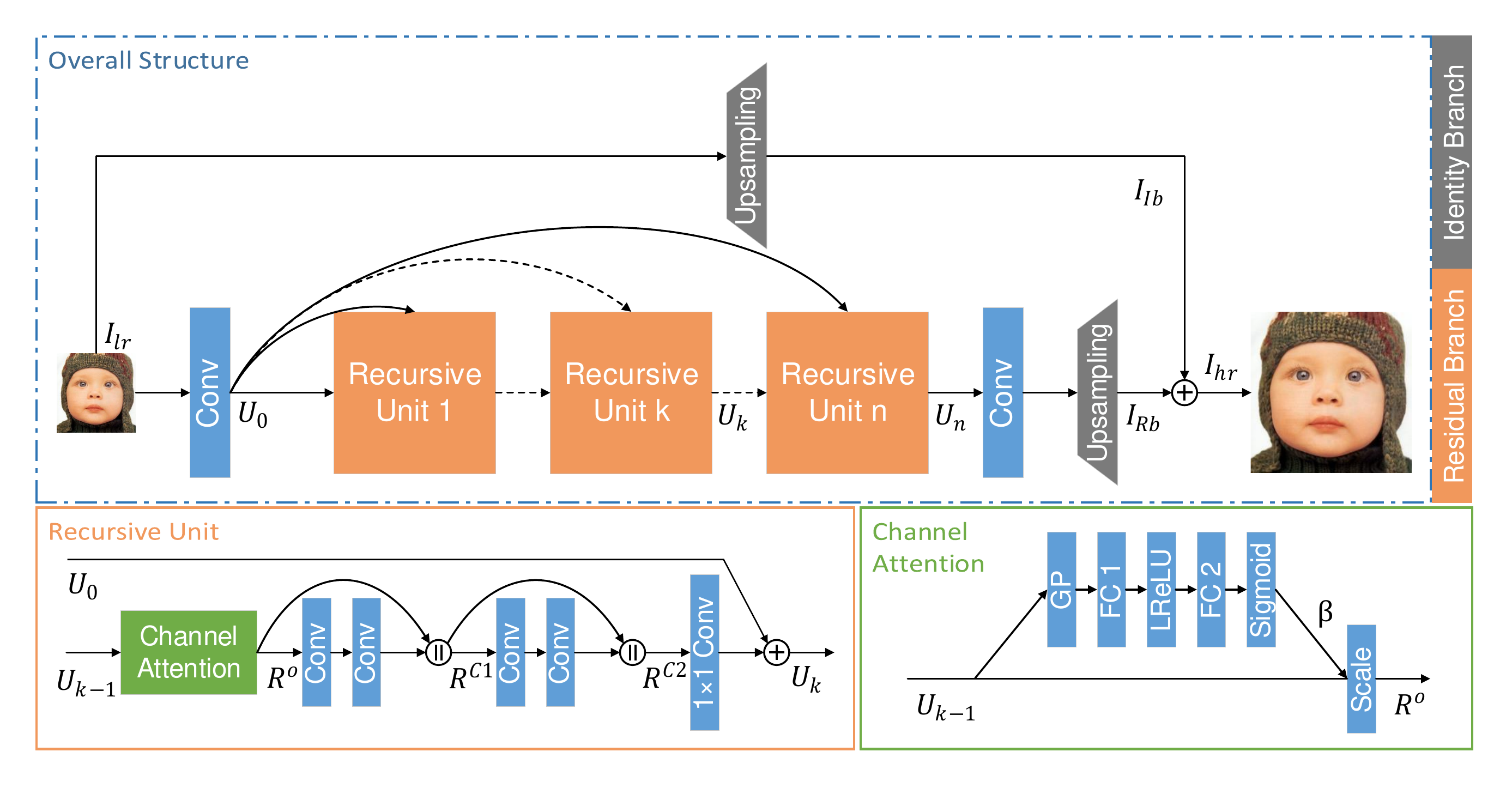}

 \caption{The architecture of our proposed network (for ${2 \times}$ SR), recursive unit and channel attention. The symbol $\parallel$ indicates concatenation and $GP$ denotes the global average pooling operation. Each convolution layer has a LReLU activation before it, and for concision, we omit it.}
  \label{fig1}
\end{figure*}

While these methods have significantly improved the performance of SISR, there remain three issues to be noticed:

First, a convolutional feature channel often corresponds to a specific functionality like texture extraction or intensity detection. Therefore, in certain recursion some feature channels are more significant than others \cite{7}. However, most of existing CNN-based SR methods treat the channel relationship equally without considering different importance or utilizing channel attention to flexibly adjust features.

Second, when features flow in a network, they can be roughly divided into original, shallow-level and deep-level features. These hierarchical features rely on different receptive fields and therefore carry diverse information. Most of existing methods, however, do not fully exploit the hierarchical features or simply combine these features in a element-wise summation manner. Despite some methods \cite{4,5,6} adopt local skip connections, their basic recursion is still a plain network (i.e., cascade of convolutional layers). The same shortage is also observed in many other computer vision tasks \cite{8,9}.

Third, LapSRN \cite{5} claims that using interpolation to upscale input images to the desired spatial resolution increases unnecessary computational cost and often results in visible reconstruction artifacts. Therefore, it employs transposed convolution both on the residual branch and identity branch, making two branches become learnable. However, we experimentally find that this strategy not only fails to explicitly learn sophisticated high-frequency residuals which contain more image details, but also fluctuates the training process.

To address these issues, we introduce a novel recursive unit into our model as shown in Fig. \ref{fig1}. Our main contribution is threefold: 
\begin{itemize}
\item	\emph{Channel Attention:} At the beginning of the recursive unit, we utilize a channel attention mechanism to adaptively recalibrate the channel importance of input features.
\item	\emph{Multi-level Features Fusion:} We extract the deep-level features and aggregate multi-level features simultaneously. Moreover, the LR feature map, extracted from first Conv layer, is added at the end of each recursive unit.
\item	\emph{Focusing more on details learning:} We use transposed convolution to upscale the residual branch and bicubic interpolation to upscale the identity branch. Experiments show that using a fixed method on one branch and modify the other branch is better than mutual adjustment.
\end{itemize}

\section{Proposed Method}
In this section, we will formulate the proposed method, including the recursive unit and overall structure.

\subsection{Recursive Unit}

The recursive unit is built upon the channel attention and multi-level features fusion. Let $\textbf{U}_{k-1} $, $\textbf{U}_k \in {\mathbb{R}^{H \times W \times C}}$ denote the input and output of $k$-th recursive unit, where $C$ represents the number of feature map channels.

\subsubsection*{\textbf{Channel Attention}}Our goal is to obtain a discriminative input feature of a certain recursion. Inspired by \cite{10}, a lightweight channel attention mechanism is introduced, which allows for global information to selectively emphasise informative features and restrain less useful ones via a one-dimensional vector ${\pmb{\beta }} = [{\beta _1},...,{\beta _i},...{\beta _C}] $. Each scalar ${\beta _i}$ represents the calibration factor of $i$-th channel. As illustrated in Fig. \ref{fig1}, we first adopt a global average pooling across spatial dimensions ${H \times W}$ to extract the global information $\pmb{\alpha} \in {\mathbb{R}^{1 \times 1 \times C}}$ from $\textbf{U}_{k-1} $. Then, it is followed by a dimension reduction layer with reduction ratio 4, a LReLu activation, a dimension increase layer and a sigmoid activation to generate ${\pmb{\beta }}$. The two computable layers are implemented by fully connected layers (i.e., ${1 \times 1}$ convolution layers). The final output of the recalibration (denoted as $\textbf{R}^{o} \in {\mathbb{R}^{H \times W \times C}} $) is acquired by rescaling the input features $\textbf{U}_{k-1} $ with ${\pmb{\beta }}$:
\begin{equation}
{\textbf{R}^o} = \pmb{\beta}  \odot {\textbf{U}_{k - 1}},
\end{equation}
where $\odot$ refers to channel-wise multiplication between the calibration factor ${\beta _i}$ and the feature channel $\textbf{u}_{k-1,i} \in {\mathbb{R}^{H \times W}}$, $i = 1,2,...C$.

\subsubsection*{\textbf{Multi-level Features Fusion}} In order to make full use of hierarchical features and further improve the information flow between layers, unlike LapSRN \cite{5} and DRRN \cite{6}, we fuse multi-level features in a recursive unit rather than only using deep-level features. As shown in Fig. \ref{fig1}, we denote $\textbf{R}^{c1} \in {\mathbb{R}^{H \times W \times {2C}}}$, $\textbf{R}^{c2} \in {\mathbb{R}^{H \times W \times {4C}}}$ as the outputs of the two concatenation operations:
\begin{equation}
{{{\bf{R}}^{c1}} = {{\bf{R}}^o}\parallel {H_1}({{\bf{R}}^o})},
\end{equation}
\begin{equation}
\label{1}
{{{\bf{R}}^{c2}} = {{\bf{R}}^{c1}}\parallel {H_2}({{\bf{R}}^{c1}}) = {{\bf{R}}^o}\parallel {H_1}({{\bf{R}}^o})\parallel {H_2}({{\bf{R}}^{c1}})},
\end{equation}
where the symbol $\parallel$ denotes concatenation and ${H_1}$, ${H_2}$ represent the first two and second two convolution operations respectively. In Eq. (\ref{1}), ${{\bf{R}}^o}$, ${H_1}({{\bf{R}}^o})$ and ${H_2}({{\bf{R}}^{c1}})$ can be considered as original, shallow-level and deep-level features which are concatenated in the channel dimension. Next, we utilize a ${1 \times 1}$ convolution layer to compress this dimension to $C$. To solve the gradient vanishing problem, the compressed feature map is finally added to the LR feature map $\textbf{U}_0 $:
\begin{equation}
{{{\bf{U}}_k} = {{H}_3}({{\bf{R}}^{c2}}) + {\bf{U}}_0},
\end{equation}
where ${{H}_3}$ refers to the function of ${1 \times 1}$ convolution layer.

\subsection{Overall Structure}

We adopt the pyramid structure to recover HR images, which is introduced by \cite{5}. Fig. \ref{fig1} only illustrates the ${2 \times }$ SR model which consists of the residual branch and the identity branch. On the residual branch, we use one convolution layer with LReLU to extract features $\textbf{U}_0 $ directly from the LR input. Then, several recursive units are stacked. Supposing there are n recursive units, the output $\textbf{U}_n $ can be obtained by
\begin{equation}
{{\bf{U}}_n} = {F_n}(...({F_2}({F_1}({{\bf{U}}_0})))...),
\end{equation}
where ${F_n}$ denotes the function of the $n$-th recursive unit. To avoid artifacts, we adopt a transposed convolution layer to up-sample the global residual image (denoted as $\textbf{I}_{Rb} $). On the identity branch, the LR input is up-sampled by the bicubic interpolation instead of a learnable method. In contrast to LapSRN that performs transposed convolution on the identity branch, our method stabilizes the training process and focuses more on image details. The up-sampled image ($\textbf{I}_{Ib} $) is finally combined with $\textbf{I}_{Rb} $ to estimate the HR image ($\textbf{I}_{hr} $) via an element-wise summation, which can be formulated as:
\begin{equation}
{{\bf{I}}_{hr}} = {F_{up1}}({{\bf{U}}_n}) + {F_{up2}}({{\bf{I}}_{lr}}) = {{\bf{I}}_{Rb}} + {{\bf{I}}_{Ib}},
\end{equation}
where ${F_{up1}}$ and ${F_{up2}}$ refer to the upsampling operations on the residual branch and identity branch respectively.

\begin{figure}[t]
\centering

\includegraphics[width=8.5cm]{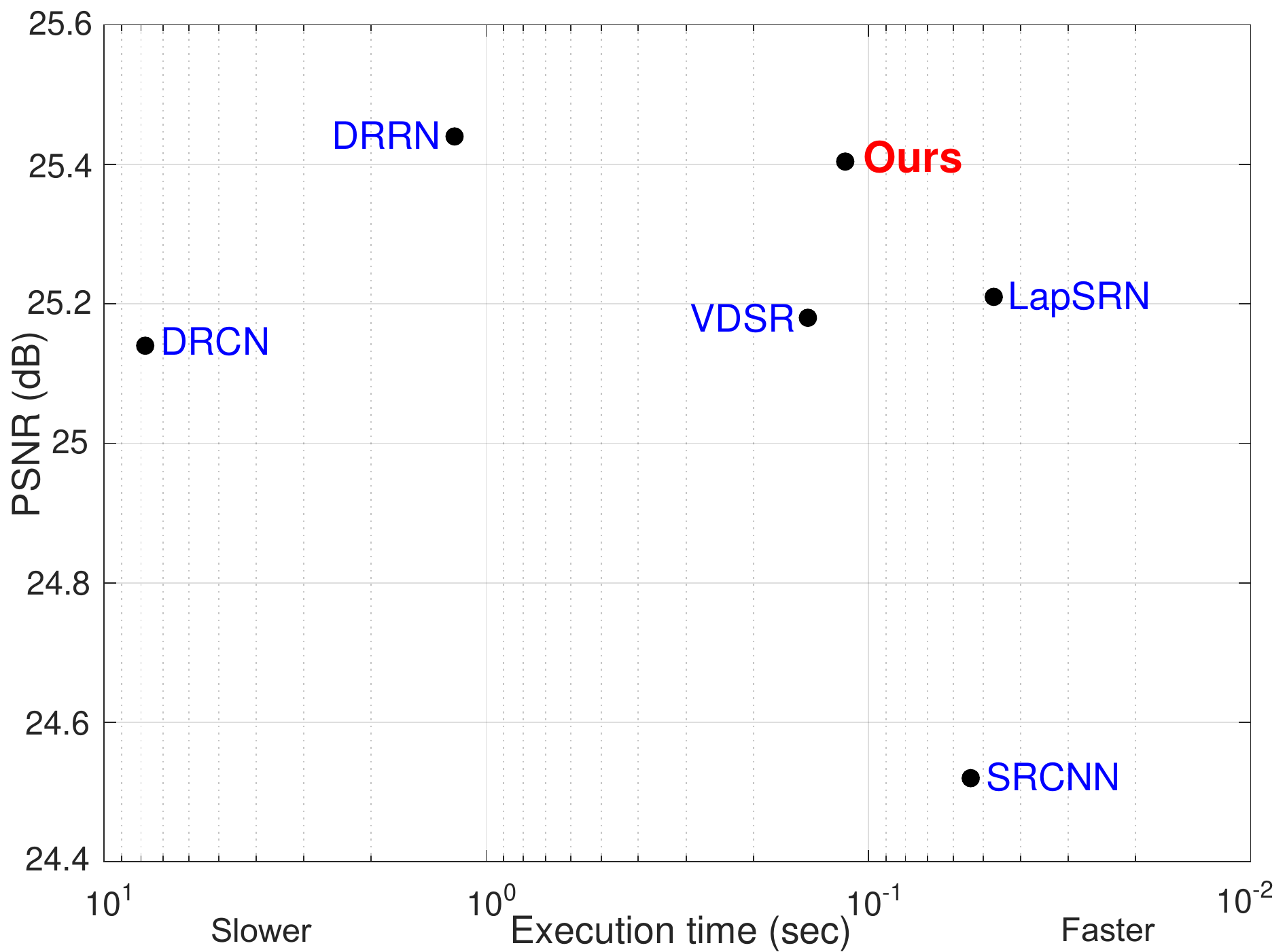}

 \caption{Runtime and performance trade-off. The results are evaluated on Urban100 \cite{16} with the scale factor ${4 \times}$.}
  \label{fig2}
\end{figure}

\begin{figure}[t]
\centering

\includegraphics[width=8.2cm]{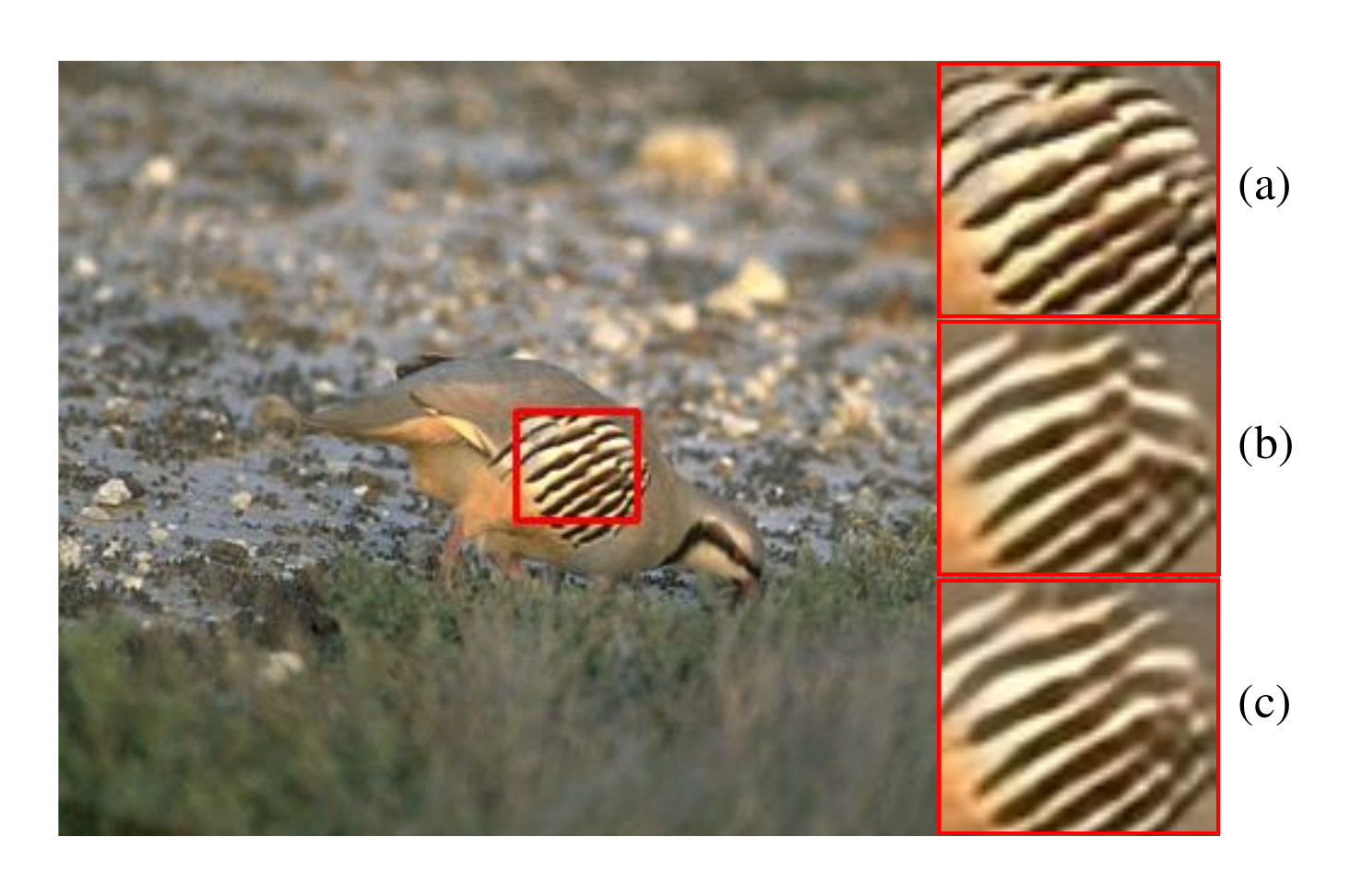}

 \caption{Reconstruction details for ${4 \times}$ SR. (a) Ground Truth. (b) Ours (using two learnable branches). (c) Ours (using one learnable branch)}
  \label{fig3}
\end{figure}

\section{Experiments}
\subsection{Implementation Details}
We train all of the models mentioned later with 291 images, where 91 images are from Yang et al. (T91) \cite{11} and other 200 images are from Berkeley Segmentation Dataset (BSDS200) \cite{12}. By following \cite{5}, we augment the training dataset via randomly scaling, rotating and flipping. For testing, we use three widely used benchmark datasets: Set5 \cite{14}, Set14 \cite{15} and BSDS100 \cite{12}. Our model is evaluated with PSNR and SSIM. We convert all images into YCbCr color space and only use the Y-channel to process.

 All recursive units share same parameters, and we use 64 convolutional filters for the first two layers and 128 filters for the second two layers. All convolutional filters (except the ${1 \times 1}$ convolution layer) have the same kernel size (${3 \times 3}$) and are initialized by the method of He et al. \cite{13}. The learning rate is set to ${10^{ - 5}}$ and decreased by a factor of 2 for every 80 epochs. We set patch size to ${128 \times 128}$ and batch size to 64. Our method is implemented with MatConvNet toolbox \cite{17} and a NVIDIA GTX 1080ti GPU.

\begin{table*}[t]
\centering
\caption{Benchmark results. Average PSNR/SSIM for scale factor ${\times 2}$, ${\times 3}$ and ${\times 4}$. Red color indicates the best performance and blue color indicates the second best performance.}
\label{table3.1}

\begin{small}
\begin{tabular}{|c|c|cc|cc|cc|cc|cc|cc|cc|}
\hline
Dataset&Scale&\multicolumn{2}{c|}{Bicubic}&\multicolumn{2}{c|}{SRCNN \cite{2}}&\multicolumn{2}{c|}{VDSR \cite{3}}&\multicolumn{2}{c|}{DRCN \cite{4}}&\multicolumn{2}{c|}{LapSRN \cite{5}}&\multicolumn{2}{c|}{DRRN \cite{6}}&\multicolumn{2}{c|}{Ours}\\
\hline
\hline
\multirow{3}{*}{Set5}
&${\times 2}$&
\multicolumn{2}{c|}{33.66/0.9299}&\multicolumn{2}{c|}{36.66/0.9542}&\multicolumn{2}{c|}{37.53/0.9587}&\multicolumn{2}{c|}{37.63/0.9588}&
\multicolumn{2}{c|}{37.52/0.9591}&\multicolumn{2}{c|}{{\color{red}37.74}/{\color{blue}0.9591}}&\multicolumn{2}{c|}{{\color{blue}37.69}/{\color{red}0.9598}}\\
&${\times 3}$&
\multicolumn{2}{c|}{30.39/0.8682}&\multicolumn{2}{c|}{32.75/0.9090}&\multicolumn{2}{c|}{33.66/0.9213}&\multicolumn{2}{c|}{33.82/0.9226}&
\multicolumn{2}{c|}{33.82/0.9227}&\multicolumn{2}{c|}{{\color{red}34.03}/{\color{red}0.9244}}&\multicolumn{2}{c|}{{\color{blue}33.94}/{\color{blue}0.9233}}\\
&${\times 4}$&
\multicolumn{2}{c|}{28.42/0.8104}&\multicolumn{2}{c|}{30.48/0.8628}&\multicolumn{2}{c|}{31.35/0.8838}&\multicolumn{2}{c|}{31.53/0.8854}&
\multicolumn{2}{c|}{31.54/0.8855}&\multicolumn{2}{c|}{{\color{red}31.68}/{\color{red}0.8888}}&\multicolumn{2}{c|}{{\color{blue}31.67}/{\color{red}0.8888}}\\
\hline
\hline

\multirow{3}{*}{Set14}
&${\times 2}$&
\multicolumn{2}{c|}{30.24/0.8688}&\multicolumn{2}{c|}{32.45/0.9067}&\multicolumn{2}{c|}{33.03/0.9124}&\multicolumn{2}{c|}{33.04/0.9118}&
\multicolumn{2}{c|}{33.08/0.9130}&\multicolumn{2}{c|}{{\color{red}33.23}/{\color{blue}0.9136}}&\multicolumn{2}{c|}{{\color{blue}33.20}/{\color{red}0.9139}}\\
&${\times 3}$&
\multicolumn{2}{c|}{27.55/0.7742}&\multicolumn{2}{c|}{29.30/0.8215}&\multicolumn{2}{c|}{29.77/0.8314}&\multicolumn{2}{c|}{29.76/0.8311}&
\multicolumn{2}{c|}{29.79/0.8320}&\multicolumn{2}{c|}{{\color{red}29.96}/{\color{red}0.8349}}&\multicolumn{2}{c|}{{\color{blue}29.89}/{\color{blue}0.8340}}\\
&${\times 4}$&
\multicolumn{2}{c|}{26.00/0.7027}&\multicolumn{2}{c|}{27.50/0.7513}&\multicolumn{2}{c|}{28.01/0.7674}&\multicolumn{2}{c|}{28.02/0.7670}&
\multicolumn{2}{c|}{28.19/0.7720}&\multicolumn{2}{c|}{{\color{red}28.21}/{\color{blue}0.7720}}&\multicolumn{2}{c|}{{\color{red}28.21}/{\color{red}0.7726}}\\
\hline
\hline

\multirow{3}{*}{BSDS100}
&${\times 2}$&
\multicolumn{2}{c|}{29.56/0.8431}&\multicolumn{2}{c|}{31.36/0.8879}&\multicolumn{2}{c|}{31.90/0.8960}&\multicolumn{2}{c|}{31.85/0.8942}&
\multicolumn{2}{c|}{31.80/0.8950}&\multicolumn{2}{c|}{{\color{red}32.05}/{\color{red}0.8973}}&\multicolumn{2}{c|}{{\color{blue}32.00}/{\color{blue}0.8971}}\\
&${\times 3}$&
\multicolumn{2}{c|}{27.21/0.7385}&\multicolumn{2}{c|}{28.41/0.7863}&\multicolumn{2}{c|}{28.82/0.7976}&\multicolumn{2}{c|}{28.80/0.7963}&
\multicolumn{2}{c|}{28.82/0.7973}&\multicolumn{2}{c|}{{\color{red}28.95}/{\color{red}0.8004}}&\multicolumn{2}{c|}{{\color{blue}28.88}/{\color{blue}0.8000}}\\
&${\times 4}$&
\multicolumn{2}{c|}{25.96/0.6675}&\multicolumn{2}{c|}{26.90/0.7101}&\multicolumn{2}{c|}{27.29/0.7251}&\multicolumn{2}{c|}{27.23/0.7233}&
\multicolumn{2}{c|}{27.32/0.7280}&\multicolumn{2}{c|}{{\color{red}27.38}/{\color{blue}0.7284}}&\multicolumn{2}{c|}{{\color{blue}27.37}/{\color{red}0.7294}}\\
\hline
%
\end{tabular}
\end{small}
\end{table*}

\subsection{Comparison with the State-of-the-Art}
Taking the trade-off between runtime and performance into account, we use 6 recursive units in our method. We compare the proposed method with 6 SR methods: Bicubic, SRCNN \cite{2}, VDSR \cite{3}, DRCN \cite{4}, LapSRN \cite{5} and DRRN \cite{6}. The quantitative results of exiting methods are quoted from their papers and shown in Table \ref{table3.1}. Our method significantly outperforms the prior methods (except DRRN) in all testing datasets and scale factors. Compared with DRRN, our method performs slightly worse on PSNR while better on SSIM in most instances, noting that SSIM focuses on measuring structural and detail similarities. Since both LapSRN and our method do not train ${3 \times }$ SR model exclusively, we achieve relatively poor performance of ${3 \times }$ SR.

As for execution time, we use the original codes of the compared methods to evaluate the runtime on a same machine with 3.6GHz Intel i7 CPU (32G RAM) and NVIDIA GTX 1080ti GPU (11G Memory). Fig. \ref{fig2} shows the trade-off between the execution time and performance on Urban100 \cite{16} dataset for ${4 \times}$ SR. Our method is 0.04dB lower than DRRN on PSNR, but approximately 10 times faster.

\subsection{Network Analysis}
To demonstrate the effect of each component, we carry out four ablation experiments of channel attention, multi-level features fusion and the number of learnable branches (i.e., using bicubic interpolation or transposed convolution in the identity branch). By removing the multi-level features fusion, our model falls back to a network similar to LapSRN \cite{5} but with the channel attention. The results confirm that making full use of multi-level features will significantly improve performance. One possible reason is that fusing hierarchical features improves the information flow and eases the difficulty of training. We can conclude from Table \ref{table3.3} that the model with full components achieves the best performance. Fig. \ref{fig3} shows the different reconstruction details between the two learnable branches and one learnable branch. Although using one learnable branch improves only 0.01dB on PSNR, it gets much better visual effects in image details.


\begin{table}[t]
\centering
\caption{Ablation experiments of our model on Set5 for ${4 \times}$ SR. Removing each component will degrade performance.}
\label{table3.3}
\begin{small}
\begin{tabular}{ccc|c}
\hline
Fusion&Attention&Learnable Branch&Set5\\
\hline
\hline
$\surd$&&One&31.62\\

&$\surd$&One&31.58\\

$\surd$&$\surd$&Two&31.66\\

$\surd$&$\surd$&One&\bf{31.67}\\

\hline
\end{tabular}
\end{small}
\end{table}


\section{Conclusion}
In this paper, a novel recursive unit is proposed for boosting single image super-resolution. The proposed unit first performs channel recalibration on input features, and then the multi-level features are extracted and fused. Moreover, we use transposed convolution on the residual branch and bicubic interpolation on the identity branch, which reconstructs more details in terms of visual perception. Experiments show that our network achieves competitive reconstruction performance and maintains faster execution speed.

\section*{Acknowledgment}

Thanks to National Natural Science Foundation of China (No.61671077 and No.61671264) for funding.

\bibliographystyle{IEEEtran}

\bibliography{IEEEluyue}
\end{document}